\documentclass{article}

\usepackage[preprint]{neurips_2023}

\usepackage[utf8]{inputenc} 
\usepackage[T1]{fontenc}    
\usepackage{hyperref}       
\usepackage{url}            
\usepackage{booktabs}       
\usepackage{amsfonts}       
\usepackage{nicefrac}       
\usepackage{microtype}      
\usepackage{xcolor}         

\usepackage{graphicx}
\usepackage{amsmath}
\usepackage{amssymb}
\usepackage{epsfig}
\usepackage{algorithm}  
\usepackage{algpseudocode}
\usepackage{pifont}
\usepackage{multirow,multicol, makecell}
\usepackage{bbm}
\usepackage{mathtools}
\usepackage{diagbox}
\usepackage{enumitem}
\usepackage{soul}
\usepackage{bbding}

\usepackage{bm}

\newcommand{\ie}{\textit{i}.\textit{e}.}
\newcommand{\eg}{\textit{e}.\textit{g}.}
\newcommand{\vs}{\textit{v}.\textit{s}.}
\newcommand{\cf}{\textit{c}.\textit{f}.}
\newcommand{\aka}{\textit{a}.\textit{k}.\textit{a}.}

\definecolor{ForestGreen}{RGB}{34,139,34}

\usepackage[capitalize]{cleveref}

\setcitestyle{square,comma,numbers}
\hypersetup{colorlinks,linkcolor={blue},citecolor={green},urlcolor={magenta}}
\definecolor{myred}{rgb}{0.75, 0.233, 0.224}
\definecolor{mygreen}{rgb}{0.482, 0.635, 0.247}
\definecolor{myblue}{rgb}{0.0, 0.361, 0.686}

\title{TreePrompt: Learning to Compose Tree Prompts for Explainable Visual Grounding}

\author{%
  Chenchi Zhang$^1$, Jun Xiao$^1$, Lei Chen$^2$, Jian Shao$^1$, Long Chen$^3$\thanks{Long Chen is the corresponding author. Work was done when Chenchi Zhang remotely visited HKUST.} \\
  $^1$Zhejiang University \; $^2$Finvolution Group \;
  $^3$Hong Kong University of Science and Technology \\
  \texttt{\{chenchiz, junx, jshao\}@zju.edu.cn, chenlei04@xinye.com,  longchen@ust.hk} \\
}

\begin{document}

\maketitle

\begin{abstract}
Prompt tuning has achieved great success in transferring the knowledge from large pretrained vision-language models into downstream tasks, and has dominated the performance on visual grounding (VG). However, almost all existing prompt tuning paradigms suffer from poor interpretability. In this paper, we argue that their poor interpretability is attributed to the holistic prompt generation and inference process. By ``holistic'', we mean that they usually directly learn a set of vectors as the prompt (\ie, prompt generation), and use the learned global prompt to augment the textual input for the VG model (\ie, prompt inference). To this end, we propose a new prompt construction paradigm with explicit explainable ability, named \emph{TreePrompt}. Specifically, we first deconstruct a complex sentence into a tree, that is consistent with human reasoning. Then, following the syntax tree, we compose a structured prompt in a bottom-up manner. Thanks to this step-by-step prompt construction process, each intermediate prompt (\ie, tree node) permits us to understand the reasoning process. Extensive ablations on various backbones and benchmarks consistently demonstrate the effectiveness and interpretability of our TreePrompt.
\end{abstract}

\section{Introduction}

Visual grounding (VG), \aka, referring expression comprehension, aims to identify and localize the referent specified by the natural language query. It is a crucial technique for many downstream applications, including navigation~\cite{chen2019touchdown}, visual question answering~\cite{antol2015vqa}, image-text matching~\cite{feng2014cross}, and so on. With the appearance of large-scale web data, the \emph{pretrain-finetune} paradigm has achieved significant success in addressing different vision-language (VL) tasks including VG, \ie, it first pretrains a general model with large-scale data, and then finetunes the model with small-size datastream data. However, fully finetuning the whole large model is always computationally expensive, and it easily suffers from over-fitting issues. Thus, in order to adopt these powerful large models to downstream tasks at a lower cost, the community recently begins to explore alternative approaches. 

\begin{figure*}[t]
\centering
\includegraphics[width=0.99\linewidth]{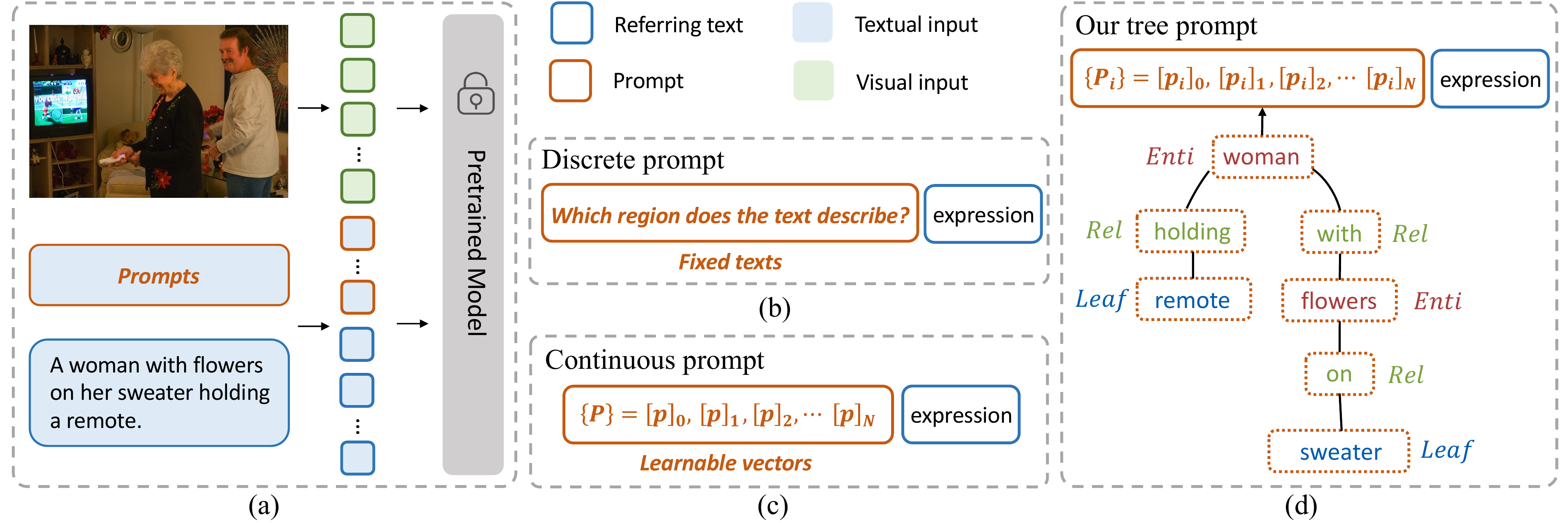}
\caption{Comparison of three types of prompt tuning methods. (a): The general framework of adapting pretrained models with prompt tuning. (b) The prompt is a manually written natural sentence (\ie, discrete prompt)~\cite{wang2022ofa}. (c) The prompt is a set of learned embeddings (\ie, continuous prompt). (d) Our proposed prompt follows the same tree structure as the query sentence.}
\label{motivation}
\end{figure*}

Among them, prompt tuning (or prompt learning)~\cite{radford2021learning} is the most popular solution for fast large model adaptation. As shown in Figure~\ref{motivation}(a), prompt tuning always fixes the parameters of the pretrained models and uses a prompt to trigger its knowledge. For example, given a well-pretrained VL model, a simple natural language prompt like ``\textit{Which region does the text describe}?'' can adapt a general VL model to the VG task (\cf, Figure~\ref{motivation}(b))~\cite{wang2022ofa}. However, manually designing suitable prompts usually requires extensive expert experience and labor. To mitigate this problem, some recent prompt tuning works try to learn the optimal prompts automatically~\cite{li2021prefix,zhou2022learning,zhou2022conditional}, \ie, the natural language prompt is substituted with a set of learnable embeddings, which are learned in an end-to-end manner.

Although significant progress has been achieved, almost all existing prompt tuning paradigms (for VG) suffer from poor interpretability in both prompt generation and inference stages: 1) In the \textbf{prompt generation} stage, they directly construct the holistic prompt without any explainable intermediate steps. 2) In the \textbf{prompt inference} stage, they usually take the global learned prompt and the query sentence as input, and then predict the corresponding output directly. In contrast, we humans typically decompose a complex reasoning task into multiple explicit subtasks. Take the query  ``\emph{A woman with flowers on her sweater holding a remote}'' in Figure~\ref{motivation} as an example, we humans would solve this task in three steps: 1) identifying the referent is ``\emph{a woman}''; 2) finding all women ``\texttt{wearing a sweater with flowers}''; and 3) localizing the woman ``\emph{holding a remote}''. Based on these observations, it is natural to raise a question: \emph{\textbf{can prompt tuning paradigm realize a similar human-like explainability}}?

In this paper, we propose a new prompt construction mechanism: \textbf{TreePrompt}, which is the first work to enable prompt-tuning-based VG methods to have explicit interpretability in both prompt generation and inference stages. Generally speaking, TreePrompt tries to compose a structured prompt following the language tree structures of query sentences, which inherently imitates the reasoning process as we humans. Specifically, we first deconstruct the complex query sentence into a tree structure with an off-the-shelf sentence parser (\eg, dependency parsing trees, DPTs~\cite{chen2014fast}). Then, we compose a structured prompt in a bottom-up manner along the tree structure. Since each word (or tree node) has different functions in localizing the referent, we design three primitive modular networks: \texttt{Leaf}, \texttt{Rel}, and \texttt{Enti} modules (detailed in Sec.~\ref{sec:modular}). Each module calculates an intermediate prompt, which simulates the human-like reasoning process step by step. For example in Figure~\ref{motivation}(d), \texttt{Rel}(\textit{holding}) combines the message carried by the word ``holding'' and prompts received from \texttt{Leaf}(\textit{remote}). Then, it will generate a new intermediate prompt that conveys the meaning like ``\emph{something is holding a remote}''. Thanks to the explicit and structured design, the intermediate prompts of different tree nodes allow us to observe the reasoning and construction process.

We evaluated our TreePrompt on three challenging VG benchmarks: RefCOCO~\cite{yu2016modeling}, RefCOCO+~\cite{yu2016modeling}, and RefCOCOg~\cite{mao2016generation}. In particular, TreePrompt not only outperforms all existing prompt tuning methods, but also achieves comparable performance to whole-model finetuning methods. Meanwhile, since TreePrompt is model-agnostic, \ie, generated prompts can be easily incorporated into different pretrained VL models, we validate its generalization ability on different architectures (\eg, OFA~\cite{wang2022ofa} and VLT5~\cite{cho2021unifying}). Extensive qualitative results have shown the strong explainable ability of TreePrompt.

In summary, we made three main contributions in this paper:
\vspace{-0.6em}
\begin{enumerate}[leftmargin=2em]
\item We propose a new TreePrompt for constructing prompts that utilize the structure of syntax trees, which effectively addresses the poor interpretability issue present in previous methods.
\item TreePrompt is universal and model-agnostic, which can be attached to any pretrained VL models.
\item  Extensive experiments have been conducted to demonstrate the superiority of our TreePrompt. 
\end{enumerate}

\section{Related Work}
\noindent\textbf{Visual Grounding.}
Existing state-of-the-art VG methods can be coarsely grouped into two categories: two-stage and one-stage methods. Specifically, \emph{two-stage methods} divide the VG task into two steps: proposal generation and proposal ranking. Firstly, a pretrained detector~\cite{girshick2014rich,girshick2015fast,ren2015faster} is used to generate proposals. Then a multimodal ranking network measures the similarity between the query sentence and proposals and selects the best result~\cite{hu2016natural,mao2016generation,rohrbach2016grounding,zhang2021vl, chen2021ref}.
This two-stage pipeline is more similar to the reasoning way of us humans. To achieve more fine-grained reasoning, some studies~\cite{yu2018mattnet,liu2019improving} decompose the query sentence and image into several components (\eg, categories, spatial relations, attributes) and solve them independently. By breaking down the complex VG task into simpler ones, each component can be analyzed and interpreted individually, which enhances models' interpretability. \emph{One-stage methods} directly predict the position of the referent~\cite{liao2020real,luo2020multi,yang2019fast}. Some studies treat VG as a conditional object detection problem~\cite{yang2019dynamic,chen2018real} and adopt an end-to-end training pipeline~\cite{zhu2022seqtr,ye2022shifting}. Although these end-to-end trainable methods may achieve better performance, their interpretability is compromised at a cost. In this paper, we propose a model-agnostic prompt generation mechanism, which can be incorporated into both one-stage and two-stage methods.

\noindent\textbf{Interpretability in Visual Grounding.}
Two-stage VG methods always achieve better interpretability. Earlier works have explored incorporating tree structures into the reasoning process for VG~\cite{liu2019learning, subramanian2022reclip,8691415}. However, they either are limited in the two-stage paradigm ~\cite{liu2019learning}, or just consider spatially related information~\cite{subramanian2022reclip}, which may not be sufficient.
More importantly, existing methods are not compatible with the prevalent prompt tuning paradigm. They rely heavily on pretrained proposal detectors which are difficult to incorporate into existing pretrained models. In contrast, our TreePrompt generates structured prompts under the prompt tuning paradigm, which inherently have strong interpretability.

\noindent\textbf{Pretrained VL Models.}
Recently, VL models~\cite{furst2022cloob,li2021supervision,lu2019vilbert,chen2020uniter,li2020oscar,gan2020large,zhang2021vinvl,wang2022ofa,lu2022unified,cho2021unifying,kamath2021mdetr} have shown powerful comprehensiveness and expressiveness, and have greatly narrowed the gap between two modalities. The success can be attributed to the development of deep networks (\eg, Transformers~\cite{vaswani2017attention}) and web-scale training datasets. By employing contrastive learning, methods like CLIP~\cite{radford2021learning} and ALIGN~\cite{jia2021scaling}, have demonstrated the ability to learn highly effective visual representations. Some methods commit to developing a unified model to work on multiple downstream tasks~\cite{lu2019vilbert,kamath2021mdetr,wang2022ofa,cho2021unifying,chen2020uniter}. They are typically pretrained on web-scale datasets, then finetuned on downstream tasks, resulting in remarkable performances in VG. In this paper, we apply some pretrained models (\eg, OFA~\cite{wang2022ofa} and VLT5~\cite{cho2021unifying}) as backbones and validate the effectiveness of our method on them.

\noindent\textbf{Prompt Tuning for VL Tasks.}
Prompt engineering involves designing appropriate instructions for guiding pretrained models in generating desired outputs. Pretrained language models such as BERT~\cite{devlin2018bert} and GPT3~\cite{brown2020language} have demonstrated impressive generalization abilities when given sophisticatedly handcrafted prompt templates. When extended to more complex multi-modal scenes, CPT~\cite{yao2021cpt} has also shown great success for zero-shot prompt learning. However, handcrafted prompts require extensive expert knowledge and experience, which limits their flexibility. In more recent works~\cite{li2021prefix, lester2021power,zhou2022learning, yang2022prompt}, researchers have proposed methods for automating prompt engineering (\eg, continuous prompts). \citet{zhou2022conditional} point out that this globally unified prompt may result in overfitting to the base class, and thus propose conditional prompt learning. However, prompt tuning suffers from severely poor interpretability. In this study, we explore a structured approach to generate prompts to enhance prompt-tuning-based VG models' interpretability.

\section{Method}

\subsection{Preliminary: Prompt Tuning in Visual Grounding}

Given an image $I$ and a textual query $T$, visual grounding (VG) aims to output the bounding box of the referent. Typically, the image is represented by a set of visual (patch/region) features, \ie, $I=\{\bm{v}_1,\cdots, \bm{v}_K\}$, where $K$ is the number of regions or patches. The query is represented by a sequence of word embeddings, \ie, $T=\{\bm{w}_1,\cdots, \bm{w}_M\}$, where $M$ is the length of word tokens and $\bm{w}_i \in \mathbb{R}^{d_t}$ denotes the embedding of $i$-th word. Currently, learning continuous prompts~\cite{zhou2022conditional} have improved VG performance and efficiency for their end-to-end trainability. Generally, continuous prompts can be classified into two types: input-layer prompt and multi-layer prompt.

\noindent\textbf{Input-Layer Prompt.} For continuous prompt, they usually introduce $N$ learnable vectors, $P = \{\bm{p}_1, \cdots, \bm{p}_N\}$, to augment the input text query, where $\bm{p}_i \in \mathbb{R}^{d_p}$ usually has the same dimension as textual embeddings $d_t$. $P$ is initialized randomly and prepended to the textual embeddings, then the augmented textual input $\hat{T}$ for the VL models becomes: 
\begin{equation}
    \hat{T}=\{\bm{v}_1, \cdots, \bm{v}_N, \bm{w}_1, \cdots, \bm{w}_M\}.
\end{equation}

\noindent\textbf{Multi-Layer Prompt.} Since most pretrained VL models apply Transformer structure as their backbones, several prompt tuning methods attempt to add prompts to each layer of the encoder and decoder. Specifically, the concatenation of prompt $P_i$, textual embeddings $T_i$ and visual embeddings $I_i$ is input to $i$-th layer of Transformer ($\mathcal{F}_i$):
\begin{equation}
    [\ \_;\ T_{i+1};\ I_{i+1}] = \mathcal{F}_i([P_{i};\ T_{i};\ I_{i}]) \quad i=1,2,\cdots, L, 
\end{equation}
where $[;]$ is a concatenation operation and $L$ is the total number of Transformer layers. $\_$ will be replaced by $P_{i+1}$ in the next layer. While multi-layer prompts improve performance, their parameters expand to $L$ times. Therefore multi-layer prompts $\mathcal{P}$ can be represented by $\mathcal{P} = \{P_1, P_2,\cdots, P_L\}$.

\begin{figure*}[t]
\centering
\includegraphics[width=0.85\linewidth]{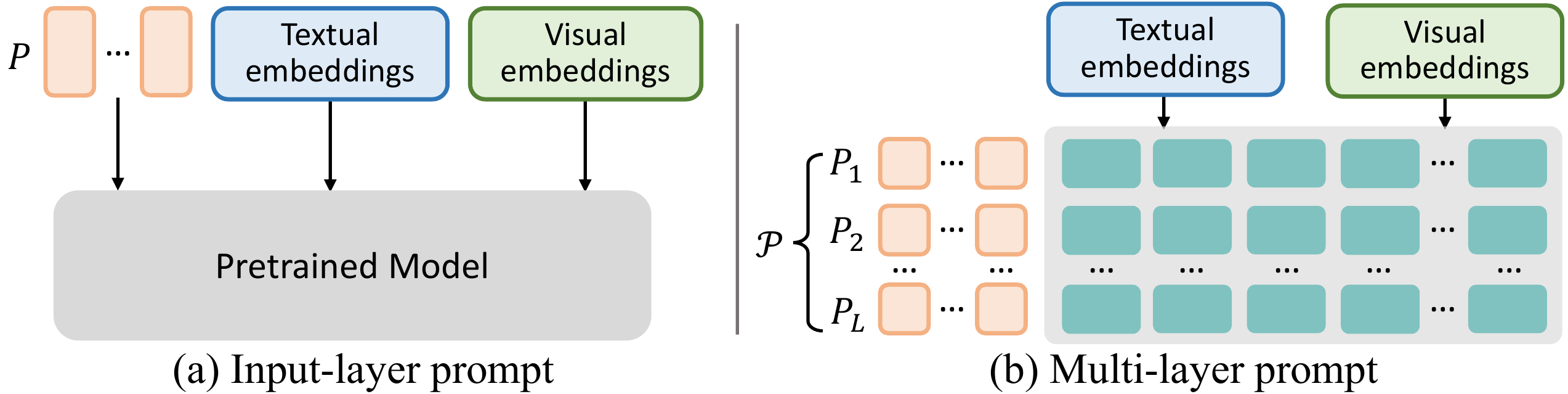}
\caption{Two types of prompts. (a)\textbf{ Input-layer prompt}: It adds the prompt tokens to the beginning of the text. (b)\textbf{ Multi-layer prompt}: It adds the prompt tokens to each layer of the Transformer.}
\label{fig:twotpye}
\end{figure*}

\noindent\textbf{Discussion.} During prompt-based training, the pretrained VL model is frozen, but the prompt can be updated through the network via backpropagation. In general, the current paradigm generates prompts in a holistic, indivisible, and unstructured manner, which we believe is the primary reason for poor interpretability. Therefore, it is essential to break down the prompt construction process into smaller, more manageable components and assemble them in a more structured, systematic manner.

\subsection{TreePrompt}

As shown in Figure~\ref{fig:method}, \textbf{TreePrompt} consists of three main steps: 1) The given sentence is parsed into a tree, where each node represents a word in the sentence. The prompt will be constructed by following the path to the tree to more effectively capture the structural relationships between the words. 2) We develop three modules (\texttt{Leaf}, \texttt{Rel}, and \texttt{Enti}) to obtain intermediate prompts for each tree node (or word). 3) The prompt generated by TreePrompt will be integrated into a global prompt (a learned prompt that is the same as the input-layer prompt).

\subsubsection{Tree Structure Generation} 
\begin{figure*}[t]
\centering
\includegraphics[width=0.99\linewidth]{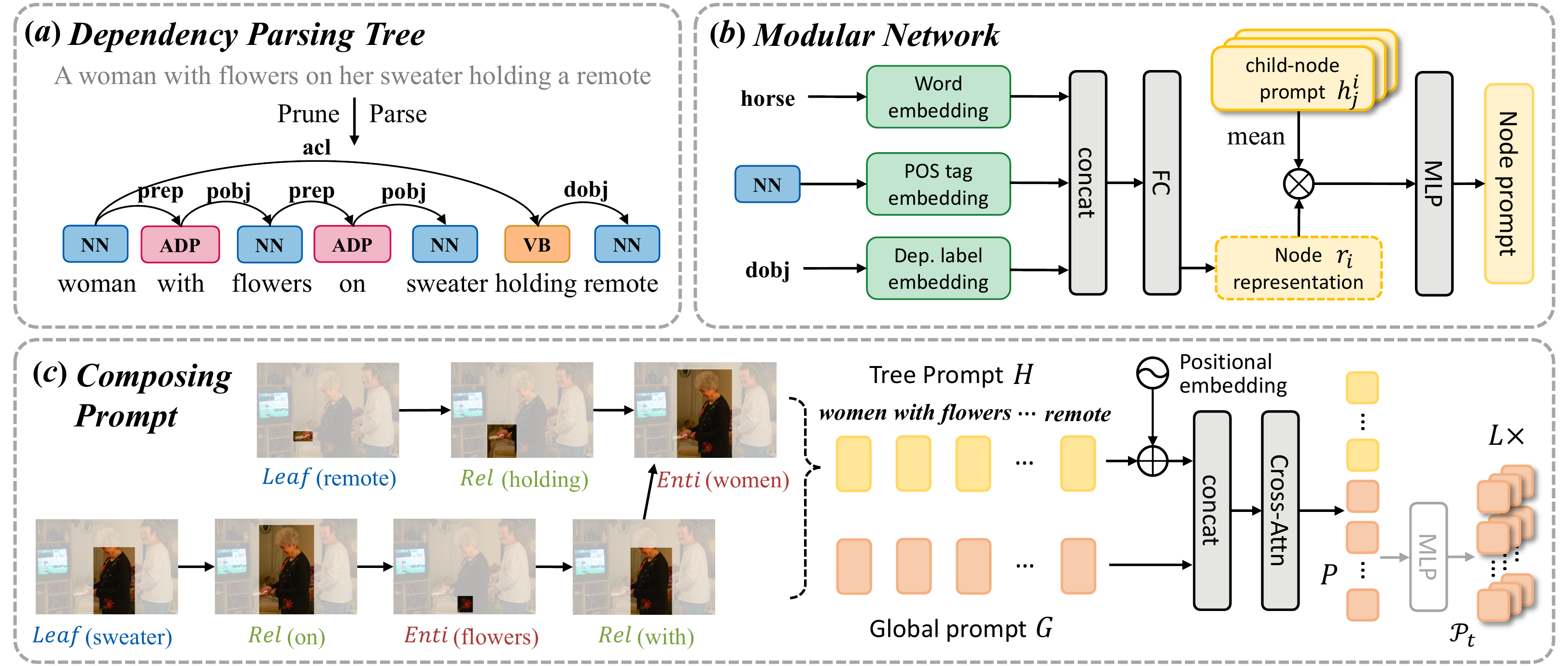}
\caption{An overview of TreePrompt model. (a) \textbf{Dependency Parsing Tree.} we first parse the input sentence into a tree structure via an off-the-shelf sentence parser. (b) \textbf{Modular Network.} we utilize it to build an intermediate prompt at each node along the tree path. (c) \textbf{Composing Prompt.} The prompts from tree are arranged in the order of traversal and combined with a global learned prompt to produce the final output. By prepending the intermediate prompt to the text and guiding the model's prediction, the reasoning process depicted in the bottom left corner can be obtained.}
\label{fig:method}
\end{figure*}

Syntax tree is the essential component in TreePrompt and provides a hierarchical representation of the sentence structure. The tree form enables us to simply identify the syntactic relationships between words, such as subject-verb-object or adjective-noun pairs. By utilizing the off-the-shelf tools of SpaCy, we can easily and quickly parse a given query sentence into a syntax tree (\cf, Figure~\ref{fig:method} (a)). The syntax tree is composed of the individual words in the sentence and each word corresponds to a node. Additionally, the DPT provides valuable information such as the part-of-speech (POS) tag $t$ and the dependency relations label $l$ between words. To fully leverage all the information, for $i$-th node, we concatenate its word embedding $\bm{w}_i \in \mathbb{R}^{d_w}$ with the corresponding POS tag $\bm{t}_i \in \mathbb{R}^{d_l}$ and dependency relation label $\bm{l}_i \in \mathbb{R}^{d_l}$, to represent the embedding of $i$-th node as $\bm{n}_i \in \mathbb{R}^{d_n}$, \ie,
\begin{equation}
\begin{gathered}
\bm{n}_i = [\bm{w}_i; \ \bm{t}_i; \ \bm{l}_i ]. \\
\end{gathered}
\end{equation}
Empirically we set $\bm{t}_i$ and $\bm{l}_i$ to the same dimension and $d_n = d_w + 2 \times d_l $. We pass the embedding of $\bm{n}_i$ to a suitable modular network (\texttt{Leaf}, \texttt{Rel}, or \texttt{Enti}) based on its dependency relation label to generate the prompt of the current node, which will be detailed in Sec.~\ref{sec:modular}.

\subsubsection{Modular Network}\label{sec:modular}
The syntax tree provides a systematic and structured pathway for constructing prompts. However, to achieve a more sophisticated reasoning approach, we design three modules (\texttt{Leaf}, \texttt{Rel}, and \texttt{Enti}) that enable the VG model to perform human-like reasoning. Each module has a distinct role and is responsible for proceeding words based on their specific characteristics and functions within the sentence. To simplify the process of module design, we intentionally use \textbf{\emph{identical}} network structures for all modules, while adjusting the parameters to fit the specific requirements of each module:
\begin{equation}
\bm{n}_i^{\prime} = \mathbf{L2Norm}(\bm{n}_i),\quad  \bm{r}_i = \mathbf{FC}(\bm{n}_i^{\prime}),
\end{equation}
where $\mathbf{L2Norm}$ is the $L_2$ normalization, and $\mathbf{FC}$ is a fully connected layer mapping $\bm{n}_i \in \mathbb{R}^{d_n}$ to $\bm{r}_i \in \mathbb{R}^{d_p}$, and $\bm{r}_i$ can be served as the representation of the node. Note $\bm{r}_i$ does not yet constitute the prompt of the current node. We construct the prompt in a bottom-up manner, following the path to the tree. Therefore, in order to obtain the prompt $\bm{h}_i$ of $i$-th node, we first obtain the representation of its child nodes. Specifically, we concatenate the representation $\bm{r}_i$ with the mean prompt embedding of its child nodes and feed it to a MLP to get the prompt of $i$-th node:
\begin{equation}
\begin{aligned}
\bm{f}_i = [\frac{\sum_{j=1}^{N_i} \bm{h}^i_j}{N_i}; \ \bm{r}_i], \quad & \bm{h}_i = \mathbf{MLP}(\bm{f}_i), \\
\end{aligned}
\end{equation}
where $\mathbf{MLP}$ represents a two-layer MLP mapping $\bm{f}_i \in \mathbb{R}^{d_p+d_p}$ to $\bm{h}_i \in \mathbb{R}^{d_p}$. For the current node $i$, $\{\bm{h}^i_j\}$ is the prompt of its child nodes and $N_i$ is the number of its child nodes. This bottom-up strategy allows us to gather relevant contextual information from surrounding words. 

\textbf{Module Functions.} Lastly, we introduce the specific functions of the three modules in detail: 1) \textbf{\texttt{Leaf} Module}: It is particularly designed for leaf nodes which differ from intermediate nodes as they have no child nodes. This module generates the initial prompt, which serves as the most fundamental and basic clue of the sentence. Eventually, these clues are aggregated along the tree path to provide a complete understanding of the sentence. 2) \textbf{\texttt{Rel} Module}: It is employed for computing the nodes whose dependency relation label is an adjectival clause (``acl'') or prepositional modifier (``prep''). These words are typically verbs or prepositions, so they tend to carry important relationships between objects in the sentence. By utilizing this module, we can obtain more precise information on the connections between various components. 3) \textbf{\texttt{Enti} Module}: It is created for the remaining nodes. These nodes usually contain information related to the entity in the sentence, such as attributes and classes. With the usage of \texttt{Enti}, we can extract more detailed information about entities and other information that may have been missed.

\textbf{Discussions}. Notably, our modular network is intentionally kept simple and relies solely on textual features for two reasons:
1) Even though the parsing tools are already highly effective, they may still contain parsing errors. Adopting simple and uniform network structures can mitigate the negative impacts of these errors.
2) Pretrained models always process visual features differently, requiring extra preprocessing steps, which can significantly restrict the generalization. By solely utilizing textual inputs, we overcome this limitation and make our method applicable to any pretrained model.

\subsubsection{Integrated with Global Prompt} \label{sec:holistic}
The tree structure and modular networks enable the generation of prompts with sentence-specific information. To make optimal use of the tree prompt, we integrate it with a global prompt that contains general knowledge about data distribution. We first arrange the prompts generated by the tree in the reverse order of traversing (similar to pre-order traversal), with the root node being the first, denoted by $H \in \mathbb{R}^{M \times d_p}$. A learnable position embedding is then attached to the tree prompt. Then, the tree prompt will be fused with the global prompt $G \in \mathbb{R}^{N \times d_p}$ through cross attention:
\begin{equation}
\begin{gathered}
    H=\{\bm{h}_1, \bm{h}_2, \cdots, \bm{h}_M\}, \quad G = \{\bm{g}_1, \bm{g}_2, \cdots, \bm{g}_N\}, \\
    [\ \_; P] = \mathbf{CrossAttn}([H; G]),
\end{gathered}
\end{equation}
where $\mathbf{CrossAttn}$ represents a cross attention network. The first vector $\bm{h}_1$ is the intermediate prompt of the root node. To apply the input-layer prompt method, $P$ is concatenated to the textual embeddings and then fed as textual input to the pretrained model. On the other hand, to implement the multi-layer prompt approach, we use an MLP to map $P \in \mathbb{R}^{N \times d_p}$ to $\mathcal{P}_t \in \mathbb{R}^{L \times N \times d_p}$, shown in Figure~\ref{fig:method} (c). $\mathcal{P}_t$ will be directly added to the global multi-layer prompt $\mathcal{P}_g \in \mathbb{R}^{L \times N \times d_p}$.


\section{Experiments}

\subsection{Experimental Settings and Details}

\noindent\textbf{Datasets.}
We validated the effectiveness of our method on three challenging VG benchmarks: 1) \textbf{RefCOCO}~\cite{yu2016modeling}, which contains 142,210 referring expressions for 50,000 objects in 19,994 images obtained through an interactive game interface. All expression-referent are split into train, val, testA, and testB sets. testA includes images with multiple people, while testB includes images with multiple objects. 2) \textbf{RefCOCO+}~\cite{yu2016modeling}, which contains 141,564 referring expressions for 49,856 objects in 19,992 images, also obtained from the same game interface and partitioned into train, val, testA, and testB splits. 3) \textbf{RefCOCOg}, which contains 104,560 referring expressions for 54,822 objects in 26,711 images obtained through a non-interactive method. We employ the same partitioning as~\cite{nagaraja2016modeling}. 

\noindent\textbf{Implementation Details and Metrics.} 
We constructed a vocabulary for each dataset, containing words, POS tags and dependency relation labels that appeared more than once in the datasets. The embedding sizes for words, POS tags, and dependency labels are set to 300, 50, and 50, respectively. We initialized the word vectors using pretrained GloVe~\cite{pennington2014glove}, while POS tags and dependency labels were initialized randomly. To obtain precise parsing results, we removed the punctuation and did not limit the expression lengths during parsing. The size of the prompt was the same as textual embeddings (\eg, 768 for OFA$_{base}$, VLT5 and 1024 for OFA$_{large}$) and the length was set to 64. We conducted experiments on 2 NVIDIA 2080Ti GPUs for VLT5 and OFA$_{base}$, and on a single NVIDIA A100 GPU for OFA$_{large}$. The weights of the pretrained model were fixed during training, while TreePrompt was trained with a batch size of 8 and a learning rate of 5e-5 via the AdamW optimizer. For OFA, we trained multi-layer prompts for 100 epochs in advance with batch size 16 and a learning rate of 0.03 following~\cite{yang2022prompt}. TreePrompt was trained with the same training objectives as the corresponding backbone models. We adopted the top-1 accuracy metric. A prediction is considered correct when the IoU between the predicted bounding box and ground truth is greater than 0.5.

\subsection{Model Agnostic Generalization}

\begin{table*}[t]
    \centering
	\caption{Performance of different models. $^\dagger$denotes the results from our implementation.}
    \scalebox{0.93}{
	\begin{tabular}{c l c c c c c c c c }
		\toprule
		\multirow{2}{*}{Models} & \multirow{2}{*}{Prompt} & \multicolumn{3}{c}{RefCOCO} &  \multicolumn{3}{c}{RefCOCO+} & \multicolumn{2}{c}{RefCOCOg} \\
		&& val & testA & testB & val & testA & testB & val & test \\
		\midrule
        \multirow{3}{*}{OFA$_{base}$} 
		& Multi-Layer Prompt$^\dagger$  & 82.28 & 86.16 & 77.02 & 75.17 & 80.61 & 65.94 & 74.88 & 75.81\\
        &~~+Continuous  & 82.50 & 86.80 & 77.17 & 74.99 & 80.82 & 66.48 & 74.43 & 75.14\\
        &~~+\textbf{TreePrompt}  & \textbf{83.36} & \textbf{87.11} & \textbf{78.19} & \textbf{75.29} & \textbf{81.05} & \textbf{66.52} & \textbf{75.12} & \textbf{76.16}\\
        \midrule
        \multirow{3}{*}{OFA$_{large}$} 
		& Multi-Layer Prompt$^\dagger$  & 89.38 & \textbf{92.45} & 84.20 & 83.24 & 89.03 & 75.76 & 84.15 & 84.81\\
        & ~~+Continuous  & 89.25 & 92.10 & 83.91 & 83.22 & \textbf{89.12} & 75.60 & 84.66 & 84.70 \\
        & ~~+\textbf{TreePrompt}  & \textbf{89.92} & 92.03 & \textbf{84.85} & \textbf{83.79} & 89.05 & \textbf{76.05} & \textbf{84.72} & \textbf{84.89}\\
        \midrule
		\multirow{2}{*}{VL-T5} 
        & ~~+Continuous  & 71.33 & 76.91 & 65.71 & 60.58 & 67.05 & 50.50 & 66.18 & 66.18  \\        
        & ~~+\textbf{TreePrompt}  & \textbf{73.96} & \textbf{79.37} & \textbf{67.81} & \textbf{62.40} & \textbf{68.88} & \textbf{51.32} & \textbf{68.10} & \textbf{68.18} \\
		\bottomrule
	\end{tabular}
    }
    \label{tab:model-agnostic}
    \vspace{-1em}
\end{table*}

\textbf{Settings}. TreePrompt is agnostic to the model architecture and can be easily integrated with any pretrained models. To evaluate the generalization, we incorporated TreePrompt into two powerful and representative pretrained methods: \textbf{OFA}~\cite{wang2022ofa} and \textbf{VLT5}~\cite{cho2021unifying}. OFA utilizes multi-layer prompts and generates object coordinates directly, while VLT5 adds prompts before the text and requires a pretrained detector to supply proposals. For OFA, we first trained the global multi-layer prompt following~\cite{yang2022prompt}, then they were fixed during training TreePrompt. We use these continuous prompts (\textbf{Continuous}) to build a strong baseline for comparison. All results are reported in Table~\ref{tab:model-agnostic}.

\textbf{Results}. TreePrompt consistently improves the performance of the multiple baselines across three benchmarks. The most significant enhancement is achieved by integrating TreePrompt with VLT5 (\eg, 2.63\%, 1.83\%, and 2.00\% absolute performance gains on three benchmarks). We believe that the difference in improvement may be caused by the multi-layer prompt used in OFA. The multi-layer prompt contains a large number of prompt parameters (\eg, with OFA$_{base}$ being 12 times larger and OFA$_{large}$ being 24 times larger than the input-layer prompt), which is sufficient for prompt tuning. Nevertheless, TreePrompt is still able to improve performance due to the benefits brought by tree structure and modular network.

\subsection{Comparison with Fully Finetuning Models}

\addtolength{\tabcolsep}{-1pt}
\begin{table*}[t]
    \caption{Comparison to state-of-the-art VG methods. The metric is top-1 accuracy(\%). $^\dagger$denotes the results from our implementations.}
    \scalebox{0.95}{
    	\begin{tabular}{l c c c c c  c cc c c c }
    		\toprule
    		\multirow{2}{*}{Models} & Finetune & \multicolumn{3}{c}{RefCOCO} & & \multicolumn{3}{c}{RefCOCO+} & & \multicolumn{2}{c}{RefCOCOg} \\
    		& entire model & val & testA & testB & & val & testA & testB & & val & test \\
    		\midrule
            UNITER~\cite{chen2020uniter} & \raisebox{-.3\height}{\includegraphics[width=2.7mm]{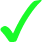}} & 81.41 & 87.04 & 74.17 & & 75.90 & 81.45 & 66.70 & & 74.86 & 75.77 \\
            VILLA~\cite{gan2020large} & \raisebox{-.3\height}{\includegraphics[width=2.7mm]{figures/true.png}} & 82.39 & 87.48 & 74.84 & & 76.17 & 81.54 & 66.84 & & 76.18 & 76.71 \\
            VLT5$^\dagger$~\cite{cho2021unifying} & \raisebox{-.3\height}{\includegraphics[width=2.7mm]{figures/true.png}} & 79.46 & 85.61 & 72.91 & & 69.17 & 76.95 & 59.62 & & 72.39 & 72.14 \\
            MDETR~\cite{kamath2021mdetr} & \raisebox{-.3\height}{\includegraphics[width=2.7mm]{figures/true.png}} & 87.51 & 90.40 & 82.67 & & 81.13 & 85.52 & 72.96 & & 83.35 & 83.31 \\
            UNICORN~\cite{yang2021crossing} & \raisebox{-.3\height}{\includegraphics[width=2.7mm]{figures/true.png}} & 88.29 & 90.42 & 83.06 & & 80.30 & 85.05 & 71.88 & & 83.44 & 83.93 \\
            PEVL~\cite{yao2022pevl} & \raisebox{-.3\height}{\includegraphics[width=2.7mm]{figures/true.png}} & 89.60 & 92.50 & 85.00 & & 83.00 & 88.40 & 74.50 & & 87.10 & 86.30 \\
            OFA$_{Large}$~\cite{wang2022ofa} & \raisebox{-.3\height}{\includegraphics[width=2.7mm]{figures/true.png}} & 90.05 & 92.93 & 85.26 & & 84.49 & 90.10 & 77.77 & & 84.54 & 85.20 \\
            ~~~~+\textbf{TreePrompt} & \raisebox{-.3\height}{\includegraphics[width=3mm]{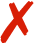}} & 89.92 & 92.03 & 84.85 & & 83.79 & 89.05 & 76.05 & & 84.72 & 84.89 \\
    		\bottomrule
    	\end{tabular}
    } 
	\label{tab:sota}
\end{table*}
\addtolength{\tabcolsep}{1pt}

\textbf{Settings}. To the best of our knowledge, most existing prompt tuning methods in VG adopt the setting of zero-shot or few-shot, which is different from our methods. Therefore, we compared our prompt tuning method with other complete model finetuning methods. As a matter of fact, such a comparison is also unfair. Finetuning the whole model involves much larger parameters than prompt tuning methods. To provide a comprehensive comparison, we present the following state-of-the-art, web-scale data pretrained VL models for comparison: \textbf{UNITER}~\cite{chen2020uniter}, \textbf{VILLA}~\cite{gan2020large}, \textbf{VLT5}~\cite{cho2021unifying}, \textbf{MDETR}~\cite{kamath2021mdetr}, \textbf{UNICORN}~\cite{yang2021crossing}, \textbf{PEVL}~\cite{yao2022pevl}, and \textbf{OFA}~\cite{wang2022ofa}.

\textbf{Results}. Although such a comparison is not fair and presents a significant challenge for TreePrompt, it still outperforms most of them, demonstrating the superiority of our TreePrompt. One may concern the improvement was solely brought by the strong backbone. We argue that TreePrompt achieves comparable performance to the powerful backbone OFA, with a much more challenging setting of only finetuning prompts. As the result shown in Table~\ref{tab:sota}, the margin is narrowed to 0.13\%, 0.7\% and -0.18\%. In general, the experimental results demonstrate the effectiveness of our TreePrompt.

\subsection{Ablation Studies}

\textbf{Effectiveness of Tree Structure and Modular Networks.} We conducted extensive ablation studies to expose the effectiveness of TreePrompt. We decomposed our method and validated each part to illustrate their functions. In Table~\ref{tab:ablation}, the term ``Tree'' indicates that the model utilizes tree structures to construct prompts, while the term ``Module'' represents the use of modular networks. We mark the use of each component (\ie, Tree and Module) with a check mark to indicate its activation and a cross mark to indicate its deactivation in the corresponding experiment.

\emph{Results}. As shown in Table~\ref{tab:ablation}, we can observe that the complete TreePrompt outperforms the incomplete version. It demonstrates the necessity of both components. The difference in the performance of the two components reveals their different roles. Compared with TreePrompt w/o Module, TreePrompt w/o Tree shows better performance in RefCOCO, while it is the opposite in RefCOCOg. We attribute it to the difference in datasets. RefCOCO has shorter and simpler expressions as compared to RefCOCOg, which has longer and more complex expressions (3.5 \vs~8.4 words per sentence on average). Modular networks are effective for precise and fine-grained reasoning, while tree structure assists in organizing sentence structure and capturing semantic information.

\begin{figure*}[t]
    \centering
    \includegraphics[width=0.99\linewidth]{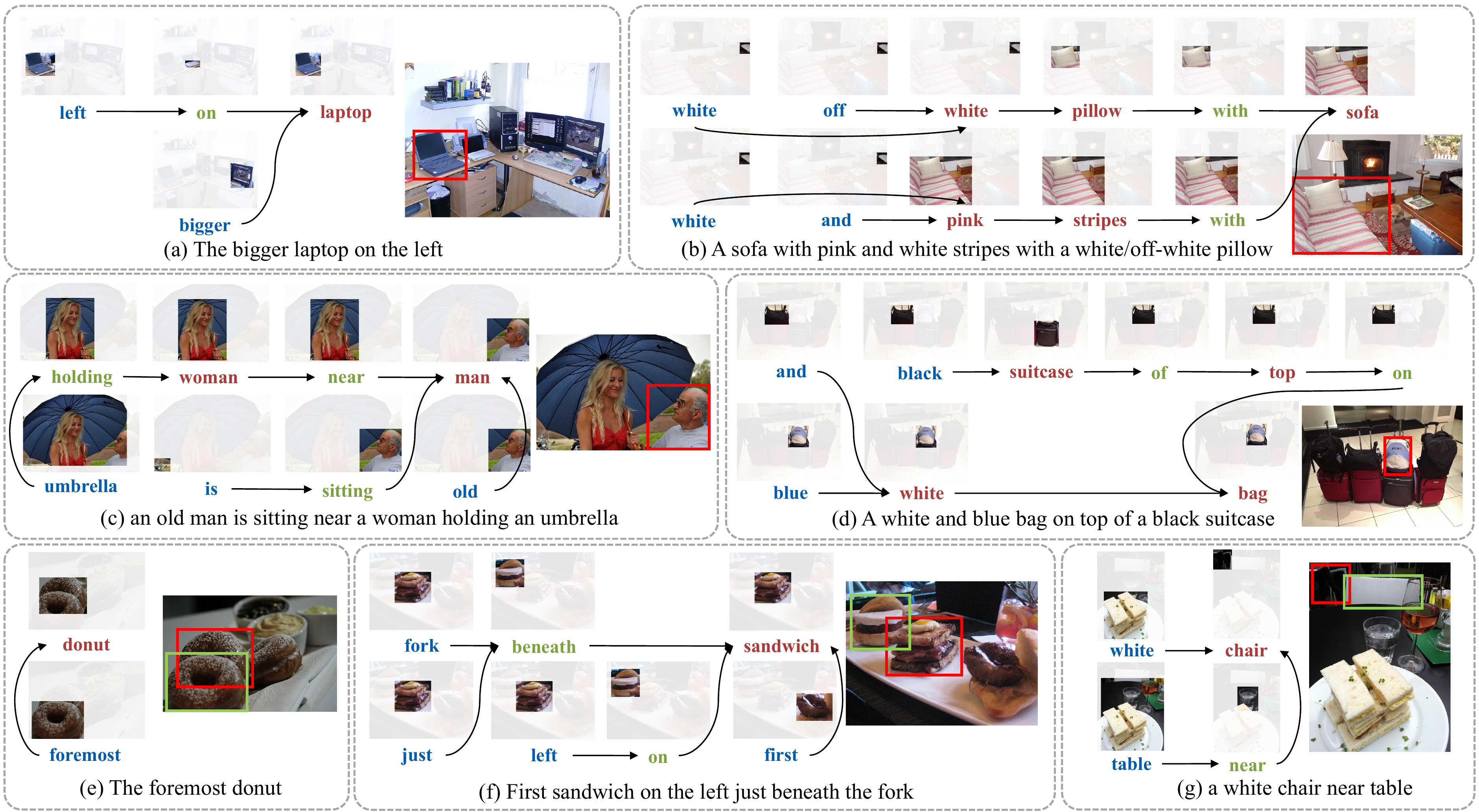}
    \caption{Qualitative results of TreePrompt on RefCOCOg. \texttt{Enti}, \texttt{Rel}, and \texttt{Leaf} modules are distinguished by their respective \textcolor{red}{red}, \textcolor{ForestGreen}{green}, and \textcolor{blue}{blue} colors of the words. GT and predictions for the original image are displayed on the right side. We present four successful cases shown in (a), (b), (c), and (d), along with three additional instances of failure for comparison, shown in (e), (f), and (g).}
    \label{fig:qualitive}
    \vspace{-0.5em}
\end{figure*}

\begin{figure}[t]
    \centering
    \includegraphics[width=0.85\linewidth]{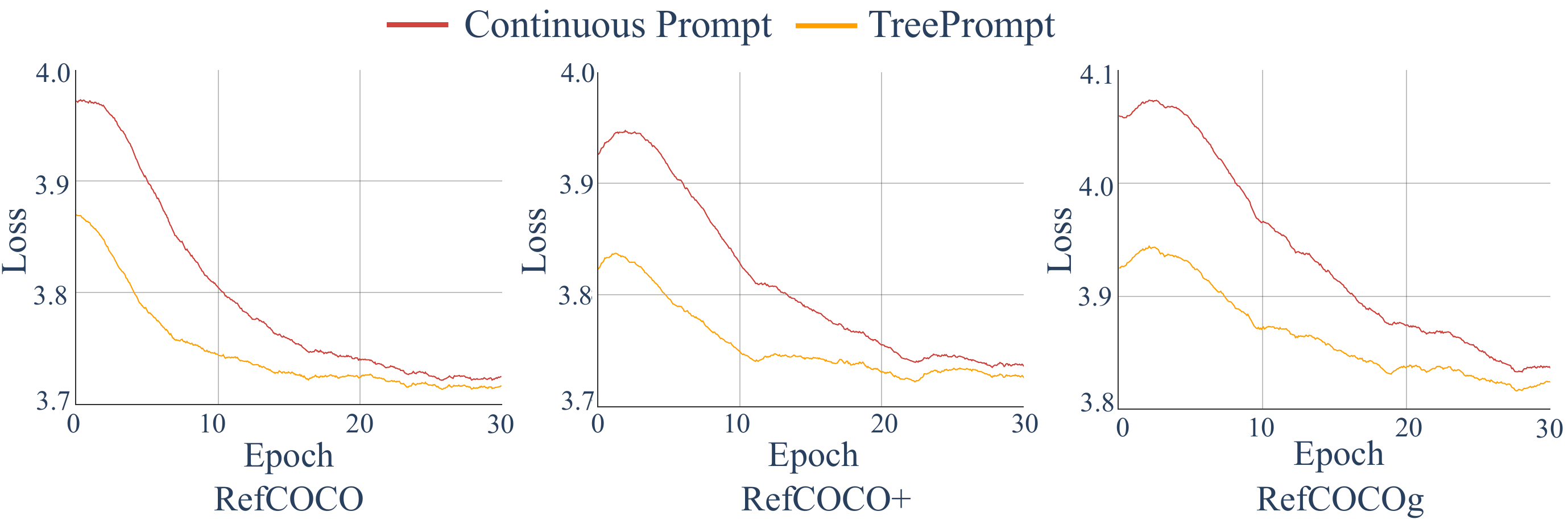}
    \vspace{-0.5em}
    \caption{Comparison of convergence rate between TreePrompt and continuous prompt on three benchmarks.}
    \label{fig:convergence}
\end{figure}

\textbf{Convergence Rate.} We compared the convergence rate of our TreePrompt and the baseline, \ie, continuous prompt. We kept the learning rate, prompt length, and other factors the same, and compared the convergence trend of the loss between the two methods. In order to comprehensively demonstrate the advantages of our method in terms of convergence rate, we conducted extensive experiments on three benchmarks (\ie, RefCOCO, RefCOCO+, and RefCOCOg). The results have shown in Figure~\ref{fig:convergence}.

\begin{table}[t]
    \begin{minipage}{0.48\linewidth}
        \addtolength{\tabcolsep}{-3pt}
        \centering
        \makeatletter\def\@captype{table}
        	\caption{Top-1 accuracy(\%) of ablation models on the three benchmarks.}
        	\scalebox{0.88}{
                \begin{tabular}{c c c c c c c c c c }
            		\toprule
            		\multirow{2}{*}{Tree} & \multirow{2}{*}{Module} & \multicolumn{2}{c}{RefCOCO} & \multicolumn{2}{c}{RefCOCO+} & \multicolumn{1}{c}{RefCOCOg} \\
            		& & testA & testB & testA & testB & test \\
            		\midrule
                    \scalebox{0.82}{\XSolidBrush} & \scalebox{0.82}{\XSolidBrush} & 
                    76.9 & 65.7 & 67.1 & 50.5 & 66.2 \\
                    \scalebox{0.82}{\XSolidBrush} & \scalebox{0.82}{\Checkmark}
                     & 
                    78.4 & 67.1 & 68.3 & 51.2 & 67.5 \\
                    \scalebox{0.82}{\Checkmark} & \scalebox{0.82}{\XSolidBrush} & 
                    78.3 & 66.7 & 68.5 & 50.5 & 67.8 \\
                    \scalebox{0.82}{\Checkmark} & \scalebox{0.82}{\Checkmark} & 
                    \textbf{79.4} & \textbf{67.8} & \textbf{68.9} & \textbf{51.3} & \textbf{68.2} \\ 
            		\bottomrule
            	\end{tabular}
            }
        \addtolength{\tabcolsep}{3pt}

    	\label{tab:ablation}
    \end{minipage}
    \hfill
    \begin{minipage}{.47\linewidth}
    \addtolength{\tabcolsep}{-2pt}
    \centering
    \makeatletter\def\@captype{table}

    	\caption{Top-1 accuracy(\%) of TreePrompt with different prompt lengths.}
        \scalebox{0.88}{
            \begin{tabular}{c c c c c c c c c }
        		\toprule
        		Prompt & \multicolumn{2}{c}{RefCOCO} & \multicolumn{2}{c}{RefCOCO+} & \multicolumn{1}{c}{RefCOCOg}  \\
        		length   & testA & testB  & testA & testB  & test  \\
        		\midrule
                32 & 78.7 & 67.5 & 67.3 & 49.8 & 65.5 \\
                64 & 79.4 & 67.8 & \textbf{68.9} & \textbf{51.3} & \textbf{68.2} \\
                100 & \textbf{79.9} & \textbf{68.6} & 67.5 & 49.4 & 65.8 \\ 
                128 & 79.5 & 68.2 & 67.8 & 49.3 &  65.2 \\ 
        		\bottomrule
        	\end{tabular}
        }
    	\label{tab:length}
    \addtolength{\tabcolsep}{2pt}
    
    \end{minipage}
\end{table}

\emph{Results}. The proposed TreePrompt decomposes the VG task into simpler components and solves them separately. This ``divide and conquer'' strategy not only leads to improved performance and fine-grained reasoning, but also speeds up convergence. As is shown in Figure~\ref{fig:convergence}, compared with the baseline, we observe that: 1) our method requires less time to reach the same loss, and 2) we can achieve a lower loss within the same time frame. These findings suggest that the model can learn more efficiently when dealing with multiple simple problems instead of a single complex one.

\textbf{Impact of Prompt Length.} Generally, prompt tuning is sensitive to the length of the prompt. Consequently, we experimented with prompts of different lengths to reveal the impact of prompt lengths on performance. Empirically, we tried 4 different lengths, \ie, 32, 64, 100, 128. For a fair comparison, all the prompt tokens were initialized randomly. The result is shown in Table~\ref{tab:length}.

\emph{Results}. In Table~\ref{tab:length}, a prompt length of 64 generally shows the best performance compared to other lengths. It is intuitive that a longer prompt can carry more information and thus boost performance. However, when the prompt length exceeds 100, the inclusion of more parameters has a negative impact on performance. This may be because having too many parameters will cause the model to overfit the training data, leading to a poor generalization of testing data.

\subsection{Qualitative Analysis}

We illustrate the qualitative results of TreePrompt in Figure~\ref{fig:qualitive}. By outputting the intermediate prompts of nodes, we are able to gain insight into the reasoning process and mechanism of TreePrompt. Through the qualitative results, we can observe that: 1) For leaf nodes that contain rich visual information, \texttt{Leaf} is able to acquire the most basic information, which provides significant assistance for reasoning, \eg, ``\emph{umbrella}'' in (c). Even for nodes containing abstract visual information, like ``\emph{bigger}'' in (a), our model can identify the biggest computer on the right. 2) \texttt{Rel} is responsible for resolving words that may involve multiple objects. It helps to identify the relationship between the objects and the action or attribute described by the word \eg, ``\emph{holding}'' and ``\emph{sitting}'' in (c). For words with ambiguous meanings like ``\emph{of}'', \texttt{Rel} may struggle to identify the object. 3) \texttt{Enti} usually has multiple subtrees, so it is capable of aggregating relevant information. 4) Along the tree path, the information will be assembled. For example, in Figure~\ref{fig:qualitive} (d), it is more likely to output the black bag because of the words ``black'' and ``top''. However, by combining the prompts generated from subtrees, the root node is able to provide more comprehensive information about the entity, leading to correct grounding results. 5) Even for the failure cases, we can still find the reason behind it by analyzing its reasoning process, which is impossible for previous unstructured prompts. 

\section{Conclusion and Limitation}
\noindent\textbf{Conclusion.} In this paper, we investigated an alternative way to construct prompts for explainable visual grounding, which has been overlooked by previous prompt tuning methods. Particularly, we drew attention to the shortcomings of the holistic and general prompt generation and inference process, which may lead to poor interpretability. To address this issue, we proposed a novel prompt construction paradigm dubbed TreePrompt, which adopts a structured manner to compose prompts. By leveraging the language syntax tree, TreePrompt constructs prompt step-by-step, and thus achieves fine-grained reasoning process and explainable results. Meanwhile, TreePrompt is compatible with various VL pretrained backbones, and can be integrated into any state-of-the-art pretrained multimodal methods. Moving forward, we aim to extend applications of TreePrompt to other multimodal tasks which confront similar interpretability challenges in prompt tuning, \eg, image captioning and VQA.

\noindent\textbf{Limitations.} Our method relies on pretrained VL models for visual grounding and thus may inherit some of the limitations and biases of those pretrained models. Another challenge is that our method requires an off-the-shelf sentence parser to generate the tree structure, which may not always be accurate or efficient. While we developed some approaches to mitigate parsing errors (detailed in Section~\ref{sec:modular}), it still could impact the accuracy of results.

\appendix

\section*{Appendix}
In the appendix, we provide more details in the following sections:
\begin{enumerate}[leftmargin=2em]
\item Broader impacts are discussed in Section~\ref{sec:broader impacts}. 
\item More detailed ablation study experiments are reported in Section~\ref{sec:ablationh}.
\item More qualitative results are shown in Section~\ref{sec:qualitative}.
\end{enumerate}

\section{Broader Impacts}\label{sec:broader impacts}

\noindent\textbf{TreePrompt.} Our method has no apparent ethical risks in terms of social negative impact and privacy violation, as all three benchmarks (\ie, RefCOCO~\cite{yu2016modeling}, RefCOCO+~\cite{yu2016modeling} and RefCOCOg~\cite{mao2016generation}) used are publicly available and transparent. 

\noindent\textbf{Pretrained backbone.} Since our method relies on pretrained multimodal models trained on web-scale data, which may unintentionally include private information, inappropriate image-text pairs, or potential biases. Given these considerations, further investigation is recommended before deploying our method in practice.

\section{Detailed Results of Ablation Studies}\label{sec:ablationh}
The detailed results of ablation studies related to modular network and prompt length are shown in Table~\ref{tab:ablation} and Table~\ref{tab:length}, respectively. 
In order to provide a comprehensive analysis, we include the performance on the validation set, which was not presented in the main paper due to space limitations. 

\noindent\textbf{Effectiveness of Tree Structure and Modular Networks.} The results are consistent with our observation in the main paper. TreePrompt w/o Tree shows superior performance in RefCOCO compared to TreePrompt w/o Module, whereas the opposite trend is observed in RefCOCOg. This suggests that modular networks contribute to precise and fine-grained reasoning, while the tree structure assists in organizing sentence structure and capturing semantic information.

\noindent\textbf{Prompt Length.} We conducted experiments using a prompt length of 10 for comparison purposes, and generally, a prompt length of 64 yielded the best performance.

\section{More Qualitative Results}\label{sec:qualitative}
\begin{figure}[t]
    \centering
    \includegraphics[width=0.99\linewidth]{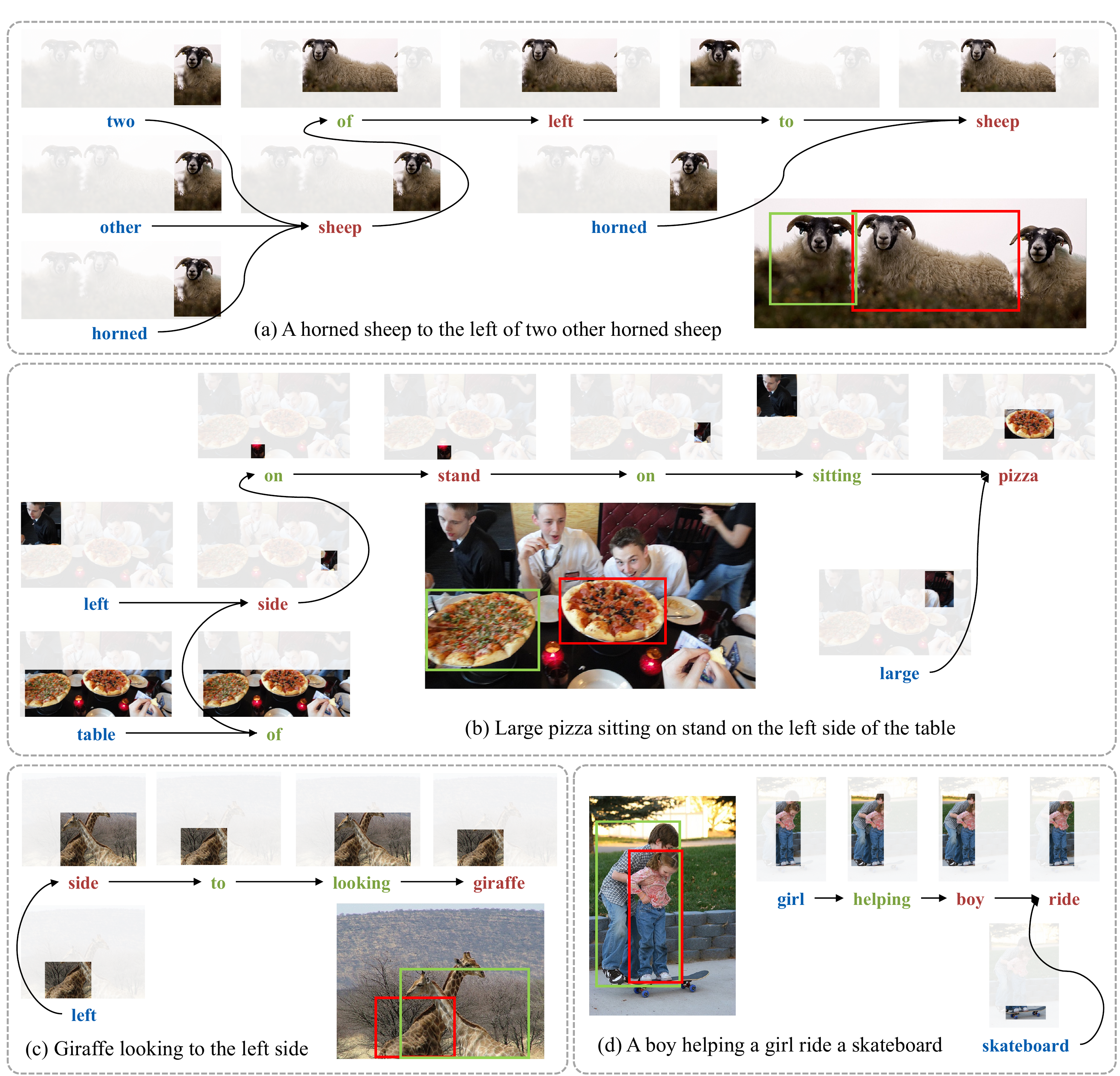}
    \vspace{-0.5em}
    \caption{More \textbf{incorrect} qualitative results of TreePrompt on RefCOCOg. \texttt{Enti}, \texttt{Rel}, and \texttt{Leaf} modules are distinguished by their respective \textcolor{myred}{red}, \textcolor{mygreen}{green}, and \textcolor{myblue}{blue} colors of the words. GT and predictions for the original image are displayed.}
    \label{fig:incorrect}
\end{figure}

\begin{figure}[t]
    \centering
    \includegraphics[width=0.99\linewidth]{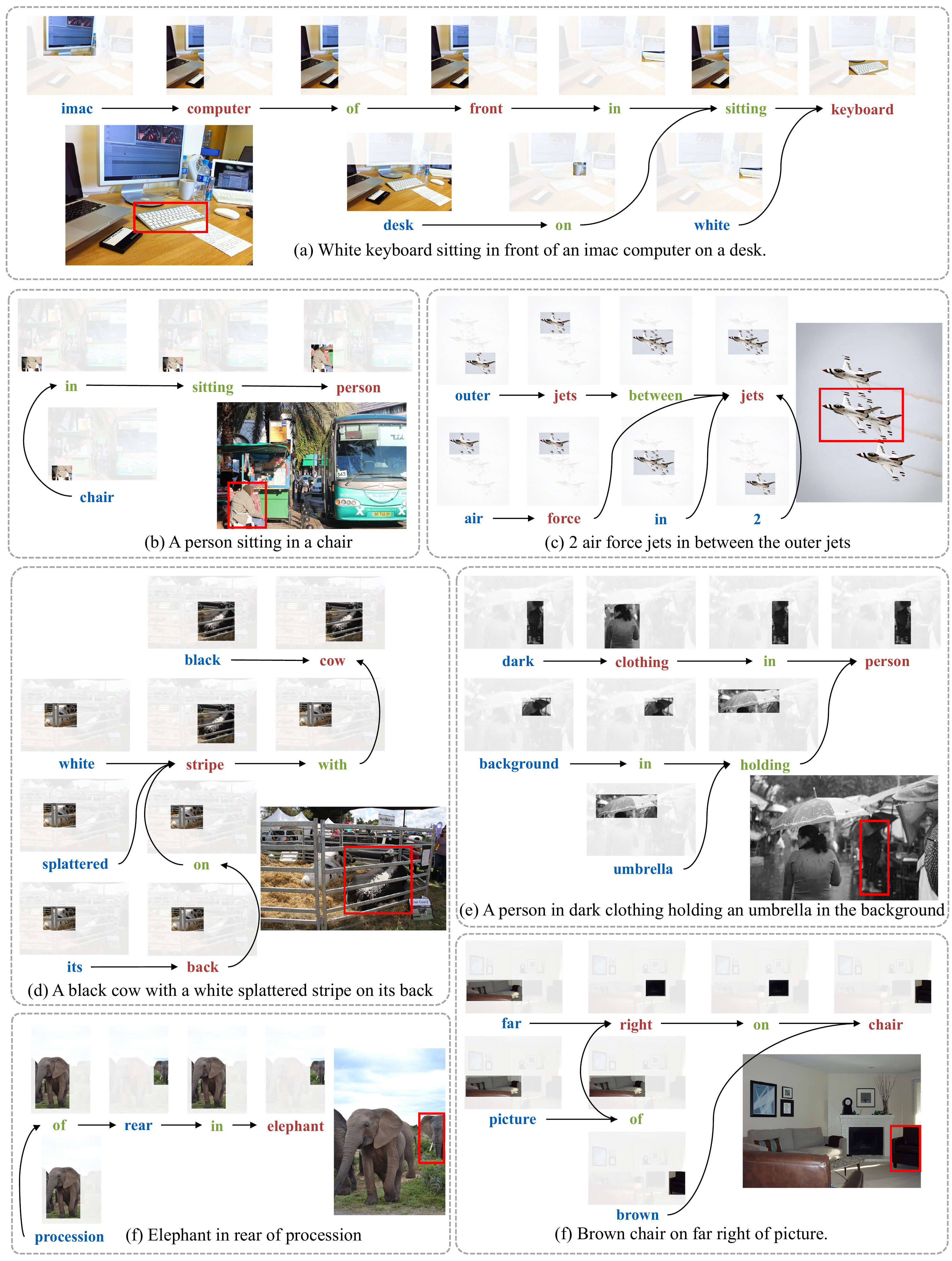}
    \vspace{-0.5em}
    \caption{More \textbf{correct} qualitative results of TreePrompt on RefCOCOg. \texttt{Enti}, \texttt{Rel}, and \texttt{Leaf} modules are distinguished by their respective \textcolor{myred}{red}, \textcolor{mygreen}{green}, and \textcolor{myblue}{blue} colors of the words. To facilitate better observation, both the original image and the ground truth (GT) are displayed.}
    \label{fig:correct}
\end{figure}

In this section, we provide additional qualitative results to better understand the internal prompt construction process employed by TreePrompt. Figure~\ref{fig:correct} shows successful cases, illustrating the tree structures, the prompt generation process at each intermediate node, and the corresponding final results. Conversely, Figure~\ref{fig:incorrect} showcases examples of failure cases for comparative analysis. Some failures arise from misunderstandings of the sentence (\eg, Figure~\ref{fig:incorrect}(a), (b), and (c)), while others are attributed to parsing errors (\eg, Figure~\ref{fig:incorrect}(d)).

\begin{table}[t]
        \centering
        	\caption{Top-1 accuracy(\%) of ablation models on the three benchmarks.}
        	\scalebox{0.99}{
                \begin{tabular}{c c c c c c c c c c c c c }
            		\toprule
            		\multirow{2}{*}{Tree} & \multirow{2}{*}{Module} & \multicolumn{3}{c}{RefCOCO} & \multicolumn{3}{c}{RefCOCO+} & \multicolumn{2}{c}{RefCOCOg} \\
            		& & val & testA & testB & val & testA & testB & val & test \\
            		\midrule
                    \scalebox{0.88}{\XSolidBrush} & \scalebox{0.88}{\XSolidBrush} & 
                    71.3 & 76.9 & 65.7 & 60.6 & 67.1 & 50.5 & 66.2 & 66.2 \\
                    \scalebox{0.88}{\XSolidBrush} & \scalebox{0.88}{\Checkmark}
                     & 
                    73.4 & 78.4 & 67.1 & 61.6 & 68.3 & 51.2 & 67.5 & 67.5 \\
                    \scalebox{0.88}{\Checkmark} & \scalebox{0.88}{\XSolidBrush} & 
                    72.7 & 78.3 & 66.7 & 61.3 & 68.5 & 50.5 & \textbf{68.3} & 67.8 \\
                    \scalebox{0.88}{\Checkmark} & \scalebox{0.88}{\Checkmark} & 
                    \textbf{74.0} & \textbf{79.4} & \textbf{67.8} & \textbf{62.4} & \textbf{68.9} & \textbf{51.3} & 68.1 & \textbf{68.2} \\ 
            		\bottomrule
            	\end{tabular}
            }
    	\label{tab:ablation}
\end{table}

\begin{table}[t]
    \centering
    \makeatletter\def\@captype{table}
    	\caption{Top-1 accuracy(\%) of TreePrompt with different prompt lengths.}
        \scalebox{0.99}{
            \begin{tabular}{c c c c c c c c c }
        		\toprule
        		Prompt & \multicolumn{3}{c}{RefCOCO} & \multicolumn{3}{c}{RefCOCO+} & \multicolumn{2}{c}{RefCOCOg}  \\
        		length & val & testA & testB & val & testA & testB & val & test  \\
        		\midrule
                10 & 70.8 & 77.0 & 64.7 & 57.7 & 65.3 & 47.7 & 63.4 & 63.7 \\
                32 & 73.0 & 78.7 & 59.3 & 67.5 & 67.3 & 49.8 & 64.9 & 65.5 \\
                64 & 74.0 & 79.4 & 67.8 & \textbf{62.4} & \textbf{68.9} & \textbf{51.3} & \textbf{68.1} & \textbf{68.2} \\
                100 & \textbf{75.0} & \textbf{79.9} & \textbf{68.6} & 59.0 & 67.5 & 49.4 & 65.5 & 65.8 \\ 
                128 & 74.3 & 79.5 & 68.2 & 59.0 & 67.8 & 49.3 & 65.4 & 65.2 \\ 
        		\bottomrule
        	\end{tabular}
        }
    	\label{tab:length}
\end{table}

\clearpage

{\small
\bibliography{egbib}
}

\end{document}